\documentclass[11pt]{article}

\usepackage[final]{acl}

\usepackage{times}
\usepackage{latexsym}

\usepackage[T1]{fontenc}

\usepackage[utf8]{inputenc}

\usepackage{microtype}

\usepackage{inconsolata}

\usepackage{graphicx}
\usepackage[percent]{overpic}

\usepackage{multirow}
\usepackage{makecell}
\usepackage{algorithm}
\usepackage{algpseudocode}
\usepackage{xcolor}  
\usepackage{tcolorbox}  
\usepackage{enumitem} 
\usepackage{tabularx} 
\usepackage{amsmath}
\usepackage{amssymb}
\usepackage{booktabs}
\usepackage{array}
\usepackage{makecell}
\usepackage{booktabs}
\usepackage{listings}

\definecolor{keywordcolor}{rgb}{0.9, 0.1, 0.1}
\definecolor{boldcolor}{rgb}{0.1, 0.1, 0.9}
\definecolor{italiccolor}{rgb}{0.1, 0.9, 0.1}

\title{SemPOI-RL: Aligning LLM Semantic Reasoning for Interpretable Out-of-Town POI Sequential Generation}

\author{
\textbf{Yunqi Liu} \quad \textbf{Yang Zhang} \quad
\textbf{Ruixing Zhang} \quad
\textbf{Liangzhe Han}\thanks{Corresponding author.} \\
\textbf{Yi Qiao} \quad \textbf{Tongyu Zhu} \quad
\textbf{Leilei Sun} \\
State Key Laboratory of Complex \& Critical Software Environment \\
Beihang University, Beijing, China \\
\texttt{\{liuyunqi, yangzhang-cn, yyxzhj, liangzhehan\}@buaa.edu.cn} \\
\texttt{\{qiaoy, zhutongyu, leileisun\}@buaa.edu.cn}
}

\begin{document}
\maketitle
\begin{abstract}
Large language models (LLMs) exhibit strong semantic reasoning and open-ended generation abilities, but aligning these abilities with structured sequential generation remains challenging. This challenge is particularly evident in out-of-town (OOT) POI sequence generation, where a model must infer transferable travel intent from a user's hometown behaviors, adapt to cross-city interest drift, and generate a coherent destination trajectory under structural constraints. Existing approaches either rely on latent ID-based transfer with limited interpretability or directly use LLMs for sequence generation without explicitly grounding inferred semantics into position-aware predictions. 
To address this gap, we propose SemPOI-RL, a framework that aligns LLM semantic reasoning with structured sequence generation for interpretable OOT recommendation. Specifically, we first fine-tune an LLM to infer destination-oriented travel styles from users' hometown trajectories, using natural language as an interpretable semantic intermediate. We then introduce a Semantic POI Alignment Module (SPAM) to ground these inferred styles into a style-conditioned masked autoencoder for position-aware trajectory generation. Finally, we apply reinforcement learning with recommendation-oriented rewards to align LLM-generated styles with downstream sequence quality. Experiments on two real-world datasets show that SemPOI-RL consistently outperforms both traditional recommenders and direct LLM baselines, while providing interpretable style attribution across different phases of a trip. The code is available at \url{https://github.com/Wind-Flipped/SemPOI-RL}.
\end{abstract}

\section{Introduction}


Large language models (LLMs) have shown impressive capabilities in semantic abstraction, contextual reasoning, and open-ended generation, making them a promising foundation for structured generation tasks \cite{TravelPlanner,TravelPlanner+,ItiNera}. However, a key challenge remains unresolved: how to align high-level semantic reasoning in natural language with downstream sequence generation that must satisfy domain-specific structural constraints \cite{LLMob,User-LLM}. This challenge is especially salient in out-of-town (OOT) POI sequence generation, where a model must infer a user's transferable travel intent from hometown behaviors and then generate a coherent destination trajectory in a new city. 
OOT POI sequence generation differs fundamentally from conventional next-POI recommendation. In standard intra-city settings, models often rely on repeated user-POI interactions, transition regularities, or region-specific spatio-temporal patterns. In contrast, the OOT setting requires the model to generalize from a user's hometown trajectory to an unfamiliar destination city, where direct POI overlap is limited and user interests may shift substantially across regions \cite{LLMMove,SemanticID, InterestDrift}. Moreover, the goal is not merely to predict the next location, but to generate an ordered sequence of destination POIs that is both semantically suitable and structurally coherent. This makes OOT recommendation a particularly difficult form of structured sequence generation under preference transfer \cite{Introduction1, LLM4POI,AR-Trip}.

\begin{figure}
 \centering
  \includegraphics[width=0.45\textwidth]{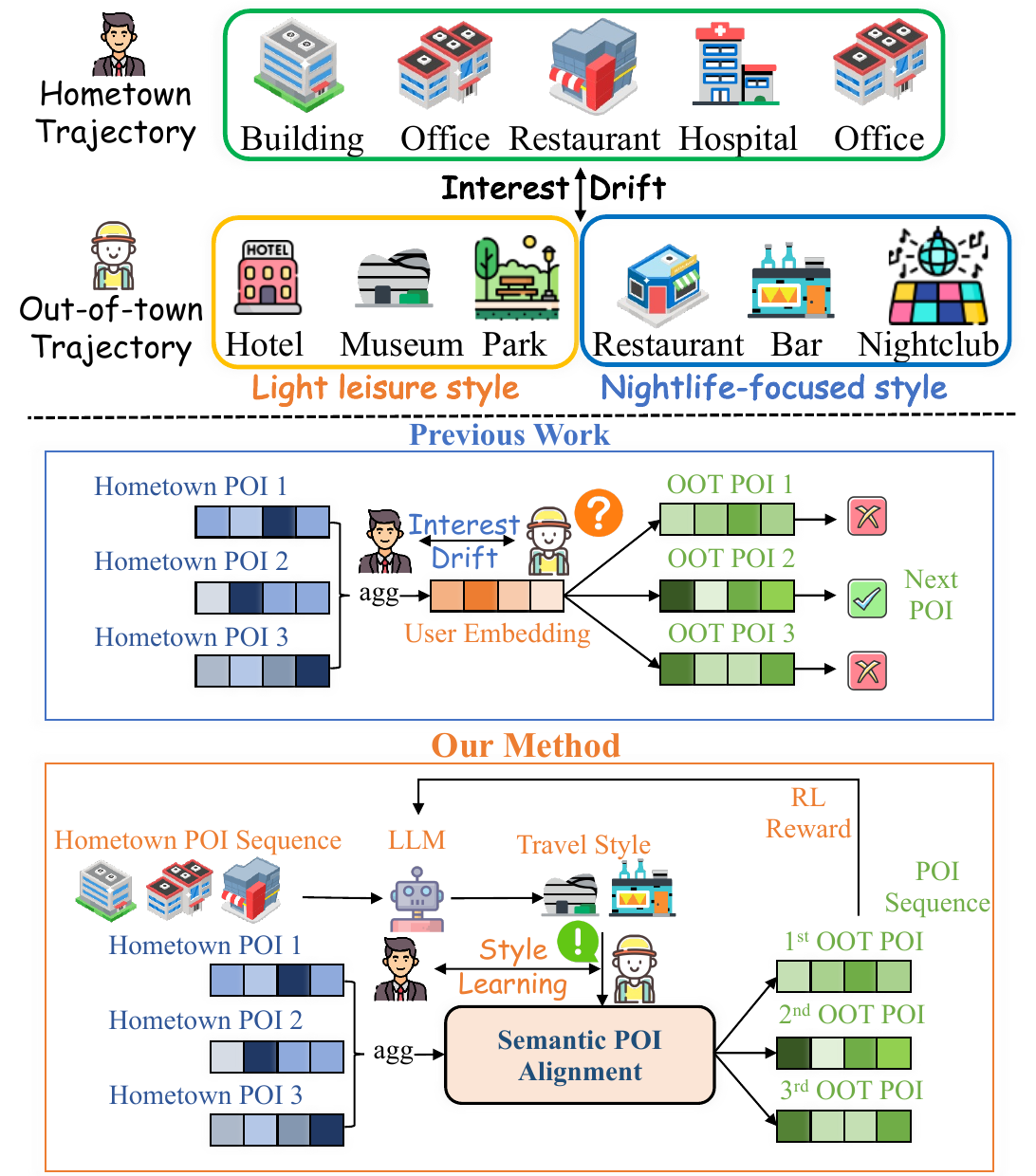}
  \caption{Comparison with previous work. Out-of-town POI sequence prediction requires the model to capture complex preferences while mitigating \emph{interest drift}. Prior approaches typically represent POIs and users independently for next-POI recommendation. Our reinforcement-learning-based framework instead aligns inferred travel styles with full-sequence prediction.}
  \label{Figure:motivation}
\end{figure}

Existing sequential recommendation and trajectory forecasting methods predominantly rely on latent ID-based representations or spatio-temporal graph encoders \cite{GETNext,STHGCN,surveyPOI}. While effective for intra-city next-POI prediction, these approaches treat sequences as flat lists of discrete tokens, offering limited interpretability and struggling to model cross-domain interest drift when users travel from familiar hometowns to novel destinations \cite{CAPTOR,SPOTTRIP}. Recent efforts have begun integrating LLMs into mobility prediction by leveraging their semantic reasoning to infer preferences or generate explanations \cite{AgentMove,POI-Enhancer,NextLocLLM}. Yet, most use the LLM as a preference encoder or direct sequence generator, map POIs to semantic IDs, or apply reinforcement learning to the final top-$k$ list \cite{User-LLM,SemanticID,LLM4POI,Refine-POI}. None of these formulations makes a natural-language travel style a trainable interface that is both aligned by trajectory rewards and decomposed across sequence positions. Consequently, the link between high-level textual inference and fine-grained structural prediction remains opaque.

To bridge these gaps, we propose \textbf{SemPOI-RL}, a reinforcement learning framework that aligns LLM semantic inference with structured sequence generation for interpretable cross-city itinerary planning. Its central contribution is a \emph{semantic--structural alignment protocol}: natural-language travel style is not merely an explanation appended after recommendation, but an intermediate representation trained from destination behavior, grounded at each trajectory position, and further optimized by the resulting POI sequence. Specifically, we first adapt an LLM via supervised fine-tuning to infer destination-oriented travel styles from users' hometown textual trajectories, effectively modeling cross-domain preference shifts through semantic reasoning \cite{CAPTOR,LLMMove}. We then introduce a \textit{Semantic POI Alignment Module} (SPAM) that injects these LLM-derived textual styles into a masked autoencoder framework. By softly assigning multiple style prototypes across sequence positions, the model exposes which semantic component guides each trip phase \cite{SemMAE}. Finally, we use Group Relative Policy Optimization (GRPO) with a composite trajectory reward to optimize the intermediate style---rather than a final top-$k$ list---for positional hit rate, recall, category consistency, and diversity \cite{GRPO,Refine-POI}.

We evaluate SemPOI-RL on two large-scale real-world datasets, demonstrating that our framework significantly outperforms both traditional sequential modeling baselines and direct LLM generation methods. Beyond quantitative gains, our approach offers substantial interpretability benefits: the learned attention maps reveal how different LLM-inferred styles dominate specific sequence positions, providing human-readable attributions for generated trajectories. This work makes three primary contributions:
\begin{itemize}
\item We propose \textbf{SemPOI-RL}, a semantic--structural alignment protocol for cross-city POI recommendation. It treats natural-language travel style as a trainable and reinforcement-aligned interface between LLM inference and structured trajectory generation, rather than as a post-hoc explanation or an opaque user embedding.

\item We design SPAM that decomposes global travel styles into temporally varying prototypical styles, and integrate it with a masked autoencoder to dynamically weight style influences across sequence positions. Furthermore, we introduce a multi-objective reinforcement learning strategy guided by hit rate, recall, category consistency, and diversity rewards, which refines LLM-generated styles to bridge the gap between semantic plausibility and structural prediction accuracy.

\item We conduct comprehensive experiments on two large-scale real-world datasets, demonstrating that SemPOI-RL consistently outperforms strong baselines in both accuracy and trajectory coherence. Detailed ablation studies and interpretability analyses further validate the necessity of each component, revealing how explicit modeling of interest drift and style-aware alignment leads to more faithful and human-understandable out-of-town recommendations.
\end{itemize}

\section{Related Work}

\subsection{POI Recommendation}

Early studies primarily framed POI sequence prediction as a Markovian process based on transition probabilities\cite{FPMC}. To address sequential dependency, models such as ST-LSTM \cite{ST-LSTM1, ST-LSTM2, MatTrip}  and transformer-based models \cite{SASRec, AR-Trip} incorporated spatio-temporal gating mechanisms to account for time intervals and geographic distances between consecutive check-ins. Meanwhile, GNNs have also gained prominence in POI recommendation due to their strong capability in modeling graph structures. For instance, GETNext\cite{GETNext} and STHGCN\cite{STHGCN} leverage GCN over trajectory flow graphs to capture POI-to-POI transition probabilities.
To tackle challenges of data sparsity and interest drift, researchers have proposed approaches based on transfer learning. \cite{CAPTOR} first studies the pre-travel out-of-town POI recommendation problem. KDDC considers the distributions of conformity and interest\cite{KDDC}. SPOT-Trip learns dual static and dynamic user preferences to enrich semantic modeling\cite{SPOTTRIP}. Traditional transfer-learning approaches typically learn mapping functions in a low-dimensional latent space. In contrast, LLMs can abstract a user's core interests into high-level, interpretable concepts without relying on specific POI IDs.

\subsection{LLMs for Recommendation}

In recent years, LLMs have emerged as a promising paradigm in recommender systems, owing to their strong natural language understanding, zero-shot and few-shot adaptability, and capacity to generate interpretable explanations\cite{TravelPlanner, TravelPlanner+}. Specifically, LLMs not only generate recommendations based on users' historical behaviors but also provide intuitive, human-readable rationales\cite{ItiNera}.
Prior work includes USER-LLM\cite{User-LLM}, which treats the LLM as a user preference encoder; and LLMob\cite{LLMob}, which infers users' occupations and travel motivations via LLMs to generate user trajectories. \cite{SemanticID} maps POI identifiers into semantically enriched representations. Moreover, LLMs have achieved strong performance on basic temporal prediction tasks\cite{TimeLLM}, inspiring extensions to next-point prediction via POI feature extraction using LLMs\cite{AgentMove, LLM-Mob, NextLocLLM, LLMMove}. Recent studies also train LLMs for POI recommendation through supervised fine-tuning\cite{LLM4POI} or reinforcement learning\cite{Refine-POI}; Refine-POI, for example, applies RL to a directly generated top-$k$ recommendation list.
However, these methods exhibit two major limitations. First, they struggle to generalize user intent across cities. Second, direct LLM decoding becomes inefficient and inaccurate when the candidate catalog exceeds the context window. More importantly, prior work does not use a natural-language travel style as a jointly trainable interface between semantic inference and position-wise trajectory generation. SemPOI-RL differs by aligning this intermediate text with ground-truth trajectory rewards and grounding it through position-specific prototype assignments; the novelty lies in this end-to-end alignment protocol rather than in any one module alone.


\begin{figure*}[t]
\centering
\begin{overpic}[width=0.95\textwidth]{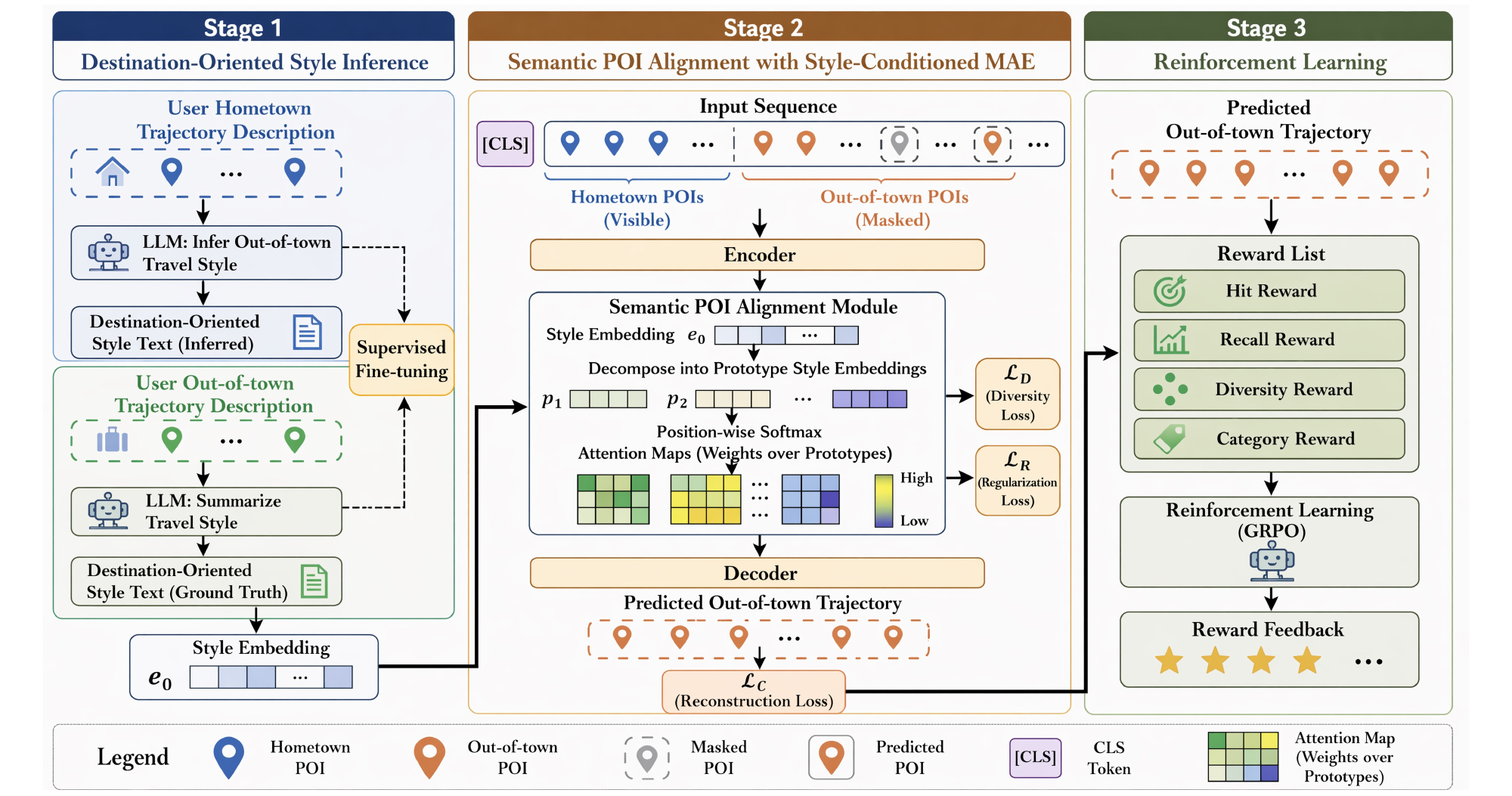}
\put(15.0,19.0){\color{white}\rule{1.2\unitlength}{4.0\unitlength}}
\end{overpic}
\caption{The overall framework of SemPOI-RL. The signal passed from Stage~1 to Stage~2 is the inferred style embedding $e_o$ generated from the hometown prompt; the destination-derived text $\mathbf{y}^{(o)}$ is used only as the Stage-1 SFT target.}
\label{figure:Model}
\end{figure*}

\section{Preliminary}
We follow prior studies\cite{CAPTOR,SPOTTRIP} to formally define the out-of-town trip recommendation problem.

\textbf{Definition 1 (POI).} 
A POI is a spatial item related to a geographical location. 
We use $v$ to represent a POI, $r_v$ to denote its located region, 
$l$ to denote its location as longitude and latitude coordinates and $g$ to denote its category.

\textbf{Definition 2 (Check-in).} 
If a user checks in a POI, it indicates that s/he has a direct interaction 
with the POI in the real world. Formally, we denote a user $u$'s check-in activity 
on POI $v$ at time $t$ with a three-tuple $c = (u, t, v)$.

\textbf{Definition 3 (User's Hometown).} 
Given a user $u$, we denote a region $r_h$ as the user's hometown 
where the user lives or works for a period of time. Following CAPTOR\cite{CAPTOR}, we identify it from time-weighted frequencies of recent check-ins after applying the same trajectory filtering described in Appendix~\ref{Appendix:datasets}.

\textbf{Definition 4 (Out-of-town Travel Behavior).} 
Given a user $u$ and her/his hometown $r_h$ in which s/he leaves check-in records 
$\vec{c}_h$, when s/he travels to the out-of-town region $r_o$ and also leaves check-in 
records $\vec{c}_o$ in $r_o$, we call such behavior an out-of-town travel behavior, 
denoted as a five-tuple $\xi = (u, \vec{c}_h, \vec{c}_o, r_h, r_o)$.

\textbf{Problem Statement: (Out-of-town Trip Recommendation).} 
Given a set of users $\mathcal{U} = \{u_i\}_{i=1}^{|\mathcal{U}|}$, 
POIs $\mathcal{V} = \{v_i\}_{i=1}^{|\mathcal{V}|}$, 
regions $\mathcal{R} = \{r_i\}_{i=1}^{|\mathcal{R}|}$, 
and out-of-town trip records $\mathcal{O} = \{\xi_i\}_{i=1}^{|\mathcal{O}|}$, 
our objective is to learn a recommender function $\mathcal{F}$ based on historical records $\mathcal{O}$. 
For a new user $u^* \notin \mathcal{U}$ at $r_h^*$ with hometown check-ins $\vec{c}_h$ 
and out-of-town origin-destination trip query 
$Q_o^* = \{v_s^o, v_e^o, N\}$ in region $r_o^*$, 
where $v_s^o$, $v_e^o$, and $N$ denote the start, end points and the number of POIs, 
the learned $\mathcal{F}$ generates a sequence of POIs $\tau = \{v_1^o, v_2^o, \ldots, v_N^o\}$,
where $v_1^o = v_s^o$ and $v_N^o = v_e^o$ for $u^*$. 
The recommended POIs in $\tau$ are in $r_o^*$. In our benchmark protocol, $u^*$ is unseen but the destination region and its POI catalog occur in the training data; thus, the structured recommender is not evaluated zero-shot on a completely new city catalog.

\section{Methodology}

We propose \textbf{SemPOI-RL}, a three-stage framework that aligns LLM-based semantic inference with structured POI sequence generation for cross-city recommendation. As discussed in the Introduction, the main challenge of out-of-town recommendation is not only to infer what a user may prefer in a new city, but also to map such high-level semantics into an ordered destination trajectory under dynamic interest drift. To address this semantic-to-structure gap, we decouple the problem into: (i) inferring destination-oriented travel styles from hometown trajectories, (ii) grounding these styles into position-aware trajectory generation, and (iii) refining style generation with recommendation-oriented rewards. Figure~\ref{figure:Model} illustrates the overall pipeline.

\subsection{Stage 1: Destination-Oriented Style Inference}
\label{sec:style_infer}

A key difficulty in cross-city recommendation is that hometown behaviors and destination behaviors are not directly aligned at the POI-ID level. Rather than transferring preferences through latent IDs only, we use an LLM to infer an interpretable \emph{destination-oriented travel style} from the user's hometown trajectory.

Given a user \(u\), hometown trajectory \(\vec c_h\) in region \(r_h\), and destination trajectory \(\vec c_o\) in region \(r_o\), we construct two prompts for an initial LLM \(\mathrm{LLM}_i(\cdot)\):
\begin{enumerate}
    \item \textbf{Inference from hometown trajectory}
    \begin{equation}
        \mathbf{y}^{(h)}=\mathrm{LLM}_i(\mathrm{Prompt}_h(u,\vec c_h,r_h)),
    \end{equation}
    where \(\mathbf{y}^{(h)}\) is the inferred destination style text.

    \item \textbf{Style summarization from destination trajectory}
    \begin{equation}
        \mathbf{y}^{(o)}=\mathrm{LLM}_i(\mathrm{Prompt}_o(u,\vec c_o,r_o)),
    \end{equation}
    where \(\mathbf{y}^{(o)}\) is the style summary directly obtained from the observed destination trajectory.
\end{enumerate}

We treat \(\mathbf{y}^{(o)}\) as a supervision signal and fine-tune the LLM to predict it from the hometown prompt. Let \(p_\theta(y_t\mid y_{<t},\mathrm{Prompt})\) denote the token probability. The supervised objective is:
\begin{equation}
\begin{aligned}
\mathcal{L}_{\mathrm{CE}}(\theta)
&= -\sum_{t=1}^{T_o} \log p_\theta \Bigl(
y^{(o)}_t \mid y^{(o)}_{<t}, \\
&\qquad \mathrm{Prompt}_h(u, \vec{c}_h, r_h)
\Bigr).
\end{aligned}
\end{equation}

After fine-tuning, the LLM generates a destination-oriented style summary:
\begin{equation}
\begin{split}
s_o&=\mathrm{LLM}_t(\mathrm{Prompt}_h(u,\vec c_h,r_h)), \\
e_o&=\mathrm{text\_embedding}(s_o),
\end{split}
\end{equation}
where $s_o$ represents the user's destination travel style, $e_o$ is the embedding of the generated style text.
A global style vector captures coarse semantic preference, but it cannot model the temporal variation emphasized in the Introduction. In practice, users often exhibit different styles at different phases of a trip (e.g., daytime leisure versus nighttime entertainment). Therefore, we further decompose the global style into multiple prototypical semantic components.

\subsection{Stage 2: Semantic POI Alignment with Style-Conditioned MAE}
\label{sec:spam_mae}

To bridge textual style inference and structured sequence prediction, we introduce the \textbf{Semantic POI Alignment Module (SPAM)} and integrate it with a masked autoencoder (MAE).

\subsubsection{Semantic POI Alignment Module}

Given the style embedding \(e_o\), SPAM decomposes it into \(M\) prototypical style embeddings:
\begin{equation}
F_p^{(i)}=
e_o \circ
\mathrm{sigmoid}\!\left(
W_{\mathrm{c2}}^{(i)}
\tanh\!\left(W_{\mathrm{c1}}^{(i)} e_o\right)
\right),
\end{equation}
where $i=1,\dots,M,$, \(\circ\) denotes the Hadamard product, and \(W_{\mathrm{c1}}^{(i)}\), \(W_{\mathrm{c2}}^{(i)}\) are learnable parameters.

This decomposition provides a set of latent style prototypes that can be selectively activated at different trajectory positions, instead of forcing the entire trip to be explained by a single semantic vector.

To avoid redundant prototypes, we impose a diversity loss:
\begin{equation}
\mathcal{L}_{D}
=
\frac{2}{M(M-1)}
\sum_{1\le i<j\le M}
\left(
\cos(\hat F_p^{(i)},\hat F_p^{(j)})
\right)^2,
\end{equation}
where \(\hat F_p^{(i)}=F_p^{(i)}/\|F_p^{(i)}\|_2\). This loss encourages the prototypes to capture distinct semantic factors.

\subsubsection{Style-Conditioned MAE}

We use the semantic prototypes to guide a masked autoencoder for destination trajectory reconstruction. For each check-in, location \(l_i\) and timestamp \(t_i\) are embedded as \(x_i=\phi(l_i,t_i)+\mathrm{pos}(i)\), where \(\phi(\cdot)\) is the input embedding function and \(\mathrm{pos}(\cdot)\) denotes positional encoding.

During training, we keep all hometown points and the destination start/end points visible and mask a proportion \(\rho\) of the intermediate destination points, yielding the partially observed sequence \(\tilde{x}\). This masking strategy matches inference, where the hometown trajectory and the destination query (start/end points and length) are given but the intermediate destination trajectory is not.

The encoder produces contextual representations \(h=E(\tilde{x})\), where \(h_i\) is the hidden state at position \(i\). To align each position with the most relevant semantic prototype, we compute a prototype distribution
\[
\alpha_i=\mathrm{SoftMax}\big(\{\langle h_i,F_p^{(k)}\rangle\}_{k=1}^{M}\big),
\]
where \(\alpha_i=[\alpha_{i,1},\dots,\alpha_{i,M}]\). This position-wise style assignment is the key step that grounds high-level style semantics into local trajectory generation.

We then form a style-aware representation \(\tilde{h}_i=\sum_{k=1}^{M}\alpha_{i,k}W_vF_p^{(k)}\), and feed it to the decoder together with the global \texttt{[CLS]} representation, i.e., \(z=D(h_{\mathrm{cls}},\tilde{h})\), where \(z\) denotes the logits over destination POIs.

The model is trained to reconstruct the masked destination positions with cross-entropy:
\begin{equation}
\mathcal{L}_{C}
=
-\frac{1}{N}
\sum_{n=1}^{N}
\log
\frac{\exp(z_{n,v_n^o})}
{\sum_{i\in r_o}\exp(z_{n,i})},
\end{equation}
where \(v_n^o\) is the ground-truth POI at the \(n\)-th masked position.
To avoid collapsing all positions to a single prototype, we maximize the normalized entropy of the style assignments. Equivalently, because the full objective is minimized, we define the following negative normalized entropy:
\begin{equation}
\mathcal{L}_{R}
=
\frac{1}{|\mathcal{I}|\log M}
\sum_{i\in\mathcal{I}}
\sum_{k=1}^{M}
\alpha_{i,k}\log \alpha_{i,k},
\end{equation}
where \(\mathcal{I}\) indexes the sequence positions. Thus, minimizing \(\mathcal{L}_R\) with a positive coefficient maximizes, rather than minimizes, assignment entropy. This is a soft anti-collapse constraint: the small default \(\lambda_R=0.1\) discourages a single prototype from dominating every position without forcing uniform assignments. Meanwhile, \(\mathcal{L}_D\) separates the prototype directions themselves, so a mildly flatter assignment still attends to genuinely distinct semantic components. Their combination avoids both globally redundant prototypes and the loss of trip-phase specificity.

The overall training objective of this section is:
\begin{equation}
\mathcal{L}
=
\mathcal{L}_{C}
+\lambda_D \mathcal{L}_{D}
+\lambda_R \mathcal{L}_{R}.
\end{equation}

\subsection{Stage 3: Reinforcement Learning for LLM Style Generation}

The supervised fine-tuning stage equips the LLM with a basic ability to infer destination-oriented travel styles from hometown trajectories, but it is still insufficient for our final objective. A key reason is that Stage~I optimizes the model only at the \emph{text} level: the LLM is trained to mimic destination-style descriptions, rather than being directly supervised by the \emph{ground-truth destination trajectory}. As a result, the generated style may be linguistically plausible yet suboptimal for downstream POI sequence prediction. We therefore introduce a reinforcement learning stage to further align style generation with the actual recommendation target. Furthermore, it relaxes the rigidity of supervised learning by allowing the LLM to explore multiple semantically reasonable style descriptions, and then retain those that lead to better structural prediction outcomes. In this way, reinforcement learning connects high-level semantic generation with downstream recommendation quality. The total reward \(R\) is defined as:
\begin{equation}
\begin{aligned}
R
&= \lambda_{HR} \cdot R_{HR}
 + \lambda_{RR} \cdot R_{RR} \\
&\quad + \lambda_{CM} \cdot R_{CM}
 + \lambda_{DIV} \cdot R_{DIV},
\end{aligned}
\end{equation}
where $R_{HR}$, $R_{RR}$, $R_{CM}$, and $R_{DIV}$ denote the hit-rate, recall, category-matching, and diversity rewards, respectively. $R_{DIV}$ is the proportion of non-repeated POIs in the prediction; the other rewards follow the metrics in Section~\ref{Exp:Metrics}. We give HR the largest weight because ordered trajectory generation most strongly requires position-level precision; RR rewards set-level coverage, CM preserves trajectory-level category semantics, and DIV prevents the repetition collapse observed in autoregressive trip generation. Because every component is computed from the predicted trajectory against the ground truth, rather than from the LLM's own textual assessment, the reward is externally grounded. In particular, DIV blocks the degenerate strategy of repeatedly predicting one popular POI. The weights are analyzed in Appendix~\ref{app:hyper_parameter}.

For policy optimization, we use GRPO\cite{GRPO}: for each hometown prompt, the policy samples a group of style descriptions, evaluates each one through the fixed trajectory predictor, and updates the LLM using group-relative rewards. This propagates recommendation quality back to the natural-language intermediate representation, providing the trajectory supervision missing from Stage~1.

\begin{table*}[!ht]
    \centering
    \footnotesize
    \setlength{\tabcolsep}{3pt}
    \begin{tabular}{l|cccccc|cccccc}
        \toprule
        \multirow{2}{*}{\textbf{Model}} & \multicolumn{6}{c|}{\textbf{Foursquare}} & \multicolumn{6}{c}{\textbf{Yelp}} \\
        \cmidrule(lr){2-7} \cmidrule(lr){8-13}
        & HR($\uparrow$) & RR($\uparrow$) & ED($\downarrow$) & DTW($\downarrow$) & CM($\uparrow$) & RM($\uparrow$) & HR($\uparrow$) & RR($\uparrow$) & ED($\downarrow$) & DTW($\downarrow$) & CM($\uparrow$) & RM($\uparrow$) \\
        \midrule
        LSTM & 0.0156 & 0.0187 & 0.9844 & 0.4920 & 0.0336 & 0.3504 & 0.0031 & 0.0032 & 0.9969 & 0.4985 & 0.0059 & 0.1255 \\
        GETNext & 0.0069 & 0.0359 & 0.9925 & 0.4966 & 0.0322 & 0.1732 & 0.0052 & 0.0108 & 0.9934 & 0.4956 & 0.0516 & 0.1379  \\ 
        STHGCN & 0.0084 & 0.0371 & 0.9910 & 0.4944 & 0.0367 & 0.2002 & 0.0066 & 0.0126 & 0.9926 & 0.4931 & 0.0587 & 0.1502 \\
        MatTrip & 0.0108 & 0.0358 & 0.9884 & 0.4942 & 0.0294 & 0.1881 & 0.0095 & 0.0301 & 0.9905 & 0.4953 & 0.0402 & 0.1760 \\
        KDDC & 0.0130 & 0.0365 & 0.9906 & 0.4931 & 0.0152 & 0.1820 & 0.0083 & 0.0329 & 0.9880 & 0.5186 & 0.0610 & 0.1045 \\
        AR-Trip & 0.0083 & 0.0308 & 0.9917 & 0.4958 & 0.0282 & \textbf{1.0000} & 0.0238 & 0.0307 & 0.9762 & 0.4880 & 0.0618 & 0.8607 \\
        SPOT-Trip & \underline{0.0190} & \underline{0.0410} & \underline{0.9806} & \underline{0.4903} & 0.0360 & \textbf{1.0000} & \underline{0.0248} & 0.0389 & \underline{0.9726} & \underline{0.4863} & \underline{0.0664} & \underline{0.9443} \\ \hline
        LLMMove & 0.0028 & 0.0355 & 0.9929 & 0.5344 & \underline{0.0388} & \underline{0.9314} & 0.0132 & 0.0256 & 0.9862 & 0.5387 & 0.0617 & 0.8843 \\
        LLM4POI & 0.0057 & 0.0083 & 0.9956 & 0.4980 & 0.0223 & 0.2036 & 0.0148 & \underline{0.0413} & 0.9836 & 0.4918 & 0.0625 & 0.3022 \\
        Refine-POI & 0.0060 & 0.0112 & 0.9940 & 0.4970 & 0.0349 & 0.3159 & 0.0065 & 0.0178 & 0.9924 & 0.4962 & 0.0484 & 0.1517 \\ \hline
        \textbf{SemPOI-RL} & \textbf{0.0219} & \textbf{0.0437} & \textbf{0.9778} & \textbf{0.4876} & \textbf{0.0572} & \textbf{1.0000} & \textbf{0.0273} & \textbf{0.0419} & \textbf{0.9701} & \textbf{0.4843} & \textbf{0.0703} & \textbf{0.9449} \\
        \bottomrule
    \end{tabular}
    \caption{The overall comparison between SemPOI-RL and baselines}
    \label{Table:overall_results}
\end{table*}

\subsection{Joint Style-Conditioned POI Sequence Prediction}

At test time, given a new user's hometown trajectory and a destination query consisting of the start POI, end POI, and trajectory length, we first use the RL-tuned LLM to infer a destination-oriented style. The ground-truth destination trajectory is never supplied as input. We then compute the prototype style embeddings with SPAM and feed them, together with the query, into the trained style-conditioned MAE to generate the intermediate destination POIs from the destination-region catalog.

\section{Experiment}

\subsection{Experiment Setup}

\paragraph{Datasets}
Our experiments are conducted on two widely used real-world datasets, \textbf{Foursquare} and \textbf{Yelp}. Detailed dataset statistics and preprocessing procedures are provided in Appendix~\ref{Appendix:datasets}.

\paragraph{Metrics}
\label{Exp:Metrics}
We evaluate all methods using six metrics:
\textbf{Hit Rate (HR)}, \textbf{Recall Rate (RR)}, \textbf{Edit Distance (ED)}, \textbf{Dynamic Time Warping Distance (DTW)}, \textbf{Category Match (CM)}, and \textbf{Region Match (RM)}.
Their formal definitions are given in Appendix~\ref{Appendix:Metrics}.

\paragraph{Settings}
Our task is more challenging than conventional next-POI prediction, because we aim to generate the \emph{entire} out-of-town POI sequence rather than predict only the next location. Following SPOT-Trip, we assume that the start and end POIs and trajectory length are given, and the model predicts the intermediate POIs. This setting reflects travel planning in which route boundaries are known but intermediate stops remain flexible; it also keeps the search space tractable for all compared methods. At test time, no other destination check-in is used as input. The train/validation/test split is 8:1:1, and Appendix~\ref{Appendix:datasets} gives a stage-by-stage data-usage protocol.

\paragraph{Baselines}
We compare SemPOI-RL with 10 baselines, including 7 traditional trip recommendation methods: LSTM~\cite{LSTM}, GETNext~\cite{GETNext}, STHGCN~\cite{STHGCN}, MatTrip~\cite{MatTrip}, AR-Trip~\cite{AR-Trip}, KDDC~\cite{KDDC}, and SPOT-Trip~\cite{SPOTTRIP}; and 3 LLM-related methods: LLMMove~\cite{LLMMove}, LLM4POI~\cite{LLM4POI}, and Refine-POI~\cite{Refine-POI}. All methods receive the same destination catalog. SemPOI-RL, SPOT-Trip, and AR-Trip hard-constrain full-sequence decoding to it; legacy next-POI architectures retain their original global output heads under our adaptation and can therefore select an out-of-region POI, which is penalized by RM. More details are provided in Appendix~\ref{Appendix:Baselines}.

\subsection{Overall Result}
In this section, we compare \textbf{SemPOI-RL} with all baselines on the two benchmark datasets. The results in Table~\ref{Table:overall_results} show that \textbf{SemPOI-RL} consistently achieves the best overall performance across both datasets. In particular, our method improves Hit Rate by about 15\% on Foursquare and 10\% on Yelp over the strongest non-LLM baselines.
Compared with traditional POI recommendation methods, the gains mainly come from explicitly modeling \emph{interest drift} between hometown and destination. Instead of relying only on latent user embeddings, SemPOI-RL uses the semantic reasoning ability of LLMs to infer destination-oriented travel styles from hometown trajectories, which provides stronger guidance for sequence generation. Compared with existing LLM-based baselines, our method also performs better under all settings, showing that simple prompting or direct fine-tuning is insufficient for this task without semantic alignment and recommendation-oriented optimization.

Among the LLM-based baselines, LLMMove shows relatively competitive Recall Rate under a simulated pre-ranked setting with 100 candidates; this protocol still assumes a strong coarse ranker capable of reducing a catalog of tens of thousands of POIs. Nevertheless, all three LLM-based baselines perform poorly on Foursquare, whose out-of-town check-ins form a smaller share of the filtered data (Appendix~\ref{Appendix:datasets}), making cross-city supervision sparser. Their lower RM under the legacy global-output adaptation further indicates difficulty preserving destination-region coherence.
By contrast, \textbf{SemPOI-RL} achieves stronger performance on \textbf{CM}, showing that it better captures category-level semantics. It also obtains lower \textbf{ED} and \textbf{DTW}, indicating that the generated trajectories are structurally closer to the ground-truth routes and require fewer edits to match real user behavior. These improvements are practically important for travel planning, where both semantic suitability and visit order matter.
Absolute exact-POI HR remains low for every method because a prediction counts only when the POI ID is correct at the exact sequence position, under a large destination catalog. CM, ED, and DTW therefore provide complementary practical evidence: an itinerary can remain useful when it chooses the right activity category and visiting order even if it misses the exact venue. SemPOI-RL leads the baselines on these measures. Three-seed results and paired tests are reported in Appendix~\ref{app:stat_reliability}.


\subsection{Ablation Study}
To validate the effectiveness of each component, we design the following ablation settings:
\begin{itemize}[leftmargin=*]
    \item \textbf{w/o SPAM}: remove the Semantic POI Alignment Module and directly fuse MAE representations with text embeddings for prediction;
    \item \textbf{w/o SFT}: skip the supervised fine-tuning stage, requiring the LLM to infer destination travel styles directly from hometown trajectories without prior adaptation;
    \item \textbf{w/o RL}: remove the reinforcement learning stage and use the style representations learned before RL for downstream POI prediction.
\end{itemize}

\begin{figure}
 \centering
  \includegraphics[width=\columnwidth]{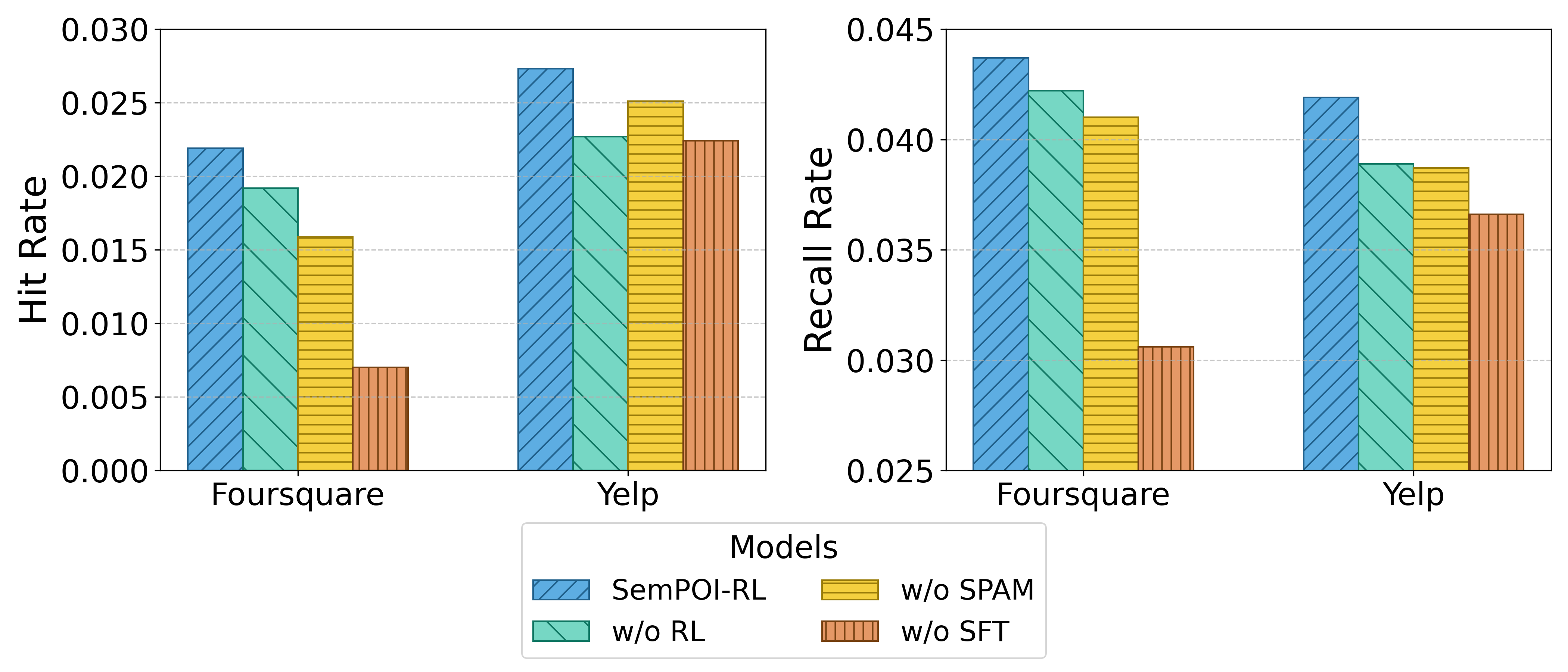}
  \caption{Ablation study of the proposed SemPOI-RL.}
  \label{Figure:Ablation}
\end{figure}

Figure~\ref{Figure:Ablation} and the quantitative results below show that all ablated variants underperform the full model across most metrics, confirming the necessity of each module.

First, removing \textbf{SFT} causes the largest performance drop, especially on Foursquare. Because this variant keeps the downstream predictor but replaces the adapted style inference with the base LLM's inference, the result directly shows that final trajectory quality depends strongly on the LLM-generated style. SFT is therefore necessary to learn the cross-city mapping rather than simply reuse hometown preference descriptions.
Second, removing \textbf{SPAM} degrades performance on both trajectory-level and semantic metrics. SPAM enables the model to decompose overall user preference into temporally varying prototype styles, which helps the generator capture different trip phases more accurately.
Third, removing \textbf{RL} also leads to consistent degradation. This shows that recommendation-oriented reward optimization further refines the LLM-inferred styles and improves their usefulness for downstream sequence generation. Additional hyper-parameter analyses are provided in Appendix~\ref{app:hyper_parameter}.

\subsection{Case Study and Interpretability Analysis}
To provide an intuitive understanding of the interpretability of \textbf{SemPOI-RL}, we present a case study in Table~\ref{Table:case_study}.
Based on the hometown trajectory, the LLM infers that the user prefers dynamic urban experiences involving recreation, exploration, and interactive activities.

As shown in Table~\ref{Table:case_study}, the predicted destination categories reveal a clear phase transition. The first part of the generated sequence mainly contains leisure, dining, and entertainment venues, whereas the latter part shifts toward rest and trip closure. This pattern is broadly consistent with the target trajectory, whose early positions are dominated by social and nightlife categories and whose later positions mainly contain accommodation-related POIs.
The prototype assignments support this interpretation: prototype \texttt{3} dominates the entertainment-oriented early positions, while prototype \texttt{1} appears in the later stadium/hotel/airport phase. The prototypes remain latent rather than deterministically nameable, so we make only a category-mediated interpretation. Full-test-set prototype/category statistics and text-similarity measurements are provided in Appendix~\ref{app:quant_interp_analysis}; additional before/after-RL cases are in Appendix~\ref{app:final_examples}.

\begin{table}[!h]
    \centering
    \scriptsize
    \begin{tabular}{m{0.22\columnwidth}|m{0.64\columnwidth}}
    \toprule
        \makecell{Hometown\\Trajectory} & [Bike Shop, Track, University, Boutique, Tech Startup] \\ \hline
        \makecell{Travel Style\\Inference} & User's travel trajectory in New York indicates a pattern of engaging with \textcolor{purple}{urban and recreational spaces}, suggesting a travel style that leans toward \textcolor{purple}{cultural and recreational exploration}. The frequent visits to a bike shop and a track suggest an interest in \textcolor{purple}{physical activity and outdoor movement}. The visit to a university points to an \textcolor{purple}{intellectual or academic curiosity}, while the boutique visit implies an appreciation for unique shopping experiences. The engagement with a tech startup further highlights an interest in innovation and modern urban environments. These behaviors collectively suggest a preference for \textcolor{purple}{dynamic, interactive, and culturally rich experiences}. The user appears to seek environments that offer both activity and discovery, favoring destinations that provide opportunities for \textcolor{purple}{exploration, learning, and engagement} with local culture and innovation. This profile aligns with a travel style characterized by curiosity, adaptability, and a desire for meaningful, experiential interactions. \\ \hline
        Predicted IDs & [159, 767, 611, 611, 186, 188, 108, 237, 6] \\ \hline
        \makecell{Predicted POI\\Category} & [\textcolor{blue}{BBQ Joint}, \textcolor{blue}{Lounge}, \textcolor{blue}{Music Venue}, \textcolor{blue}{Music Venue}, \textcolor{blue}{Rock Club}, \textcolor{blue}{Hotel Bar}, \textcolor{red}{Baseball Stadium}, \textcolor{red}{Hotel}, \textcolor{red}{Airport}] \\ \hline
        Prototype Style & [\textcolor{blue}{3}, \textcolor{blue}{3}, \textcolor{blue}{3}, \textcolor{blue}{3}, \textcolor{blue}{3}, \textcolor{blue}{3}, \textcolor{red}{1}, \textcolor{red}{1}, \textcolor{red}{1}] \\ \hline
        Target POI IDs & [159, 188, 767, 186, 239, 188, 67, 774, 1112] \\ \hline
        \makecell{Target POI\\Category} & [\textcolor{blue}{BBQ Joint}, \textcolor{blue}{Rock Club}, \textcolor{blue}{Lounge}, \textcolor{blue}{Rock Club}, \textcolor{blue}{Hotel Bar}, \textcolor{blue}{Rock Club}, \textcolor{red}{Hotel}, \textcolor{red}{Hotel}, \textcolor{red}{Hotel}] \\
        \bottomrule
    \end{tabular}
    \caption{A case study for user 2842 on the Foursquare dataset. Purple highlights key semantic cues in the inferred travel style. Blue and red indicate two dominant trip phases.}
    \label{Table:case_study}
\end{table}

\section{Conclusion}

We presented \textbf{SemPOI-RL}, a semantic-aligned framework for cross-city POI sequence recommendation. Our method addresses a core challenge of out-of-town recommendation: bridging the gap between high-level user preference semantics and structured destination trajectory generation. To this end, we first use an LLM to infer destination travel styles from users' hometown trajectories, providing an interpretable semantic abstraction of cross-region interest drift. We then introduce a Semantic POI Alignment Module to decompose global style signals into temporally varying prototypes and inject them into a style-conditioned masked autoencoder for position-aware POI sequence prediction. Finally, we further refine LLM style generation with multi-objective reinforcement learning for downstream recommendation accuracy.
Experiments on two real-world datasets show that SemPOI-RL consistently outperforms both traditional sequential recommendation models and recent LLM-based baselines. In addition to improving trajectory accuracy and structural coherence, our framework offers interpretable style attribution across different phases of a trip, providing a more transparent view of how semantic reasoning can guide sequential recommendation. We hope this work can encourage future research on aligning LLM-based semantic inference with structured decision-making tasks beyond POI recommendation.

\section*{Limitations}
SemPOI-RL has three main limitations. First, its prototypes remain latent and can only be interpreted indirectly through activated POI categories; style quality also depends on LLM-generated pseudo-labels. Second, we follow the constrained benchmark setting with known start/end POIs and trajectory length, and the recommender is trained on fixed destination catalogs rather than evaluated zero-shot in unseen cities. Open-ended generation and catalog transfer remain future work. Third, the multi-stage pipeline is costlier than simple recommenders, and exact-POI HR remains low; real-world deployment therefore requires further reliability evaluation and user studies.

\section*{Ethical Considerations}
POI trajectories are sensitive because they may reveal users' routines and private attributes. During data collection and preprocessing, all user identifiers were anonymized before the data were used for analysis. We use only structured check-in records and no user-authored free text, so offensive textual content is not involved. We do not attempt to recover user identities and retain only the fields needed for this research; any deployment should likewise apply data minimization and access controls. The framework may also inherit popularity and regional biases from check-ins, as well as stereotypes from LLM-generated styles, leading to uneven recommendation quality across users and locations.

\section*{Acknowledgments}
This work was supported by the National Natural Science Foundation of China (No. U24B20171). Generative AI tools were used to assist with the implementation of portions of the code and with grammar checking and language polishing of the manuscript. All AI-assisted outputs were reviewed and verified by the authors, who take full responsibility for the final code and manuscript.

\bibliography{sample-base}

\appendix

\section{Datasets}
\label{Appendix:datasets}
Our experiments are conducted on two widely used travel behavior datasets: \textbf{Foursquare} and \textbf{Yelp}. Following prior work\cite{CAPTOR, SPOTTRIP}, we identify users who have check-in activity both in their hometown and in out-of-town locations, and reorganize their records into cross-region travel trajectories. Hometowns are discovered with CAPTOR's time-weighted recent check-in frequencies and the filtering below. We remove users with fewer than three destination check-ins or trips shorter than 1 hour or longer than 30 days; remove POIs visited fewer than twice; and retain records satisfying $\lvert \vec{c}_h \rvert \ge 4$, $\lvert \vec{c}_o \rvert \ge 3$, and $\operatorname{freq}(r_h,r_o)\ge 10$. Stage-1 pseudo-labels are generated only from these filtered destination trajectories. A stronger teacher could further improve pseudo-label quality, but evaluating that choice is left to future work. Finally, we split users into training/validation/test sets in an 8:1:1 ratio. Table~\ref{Table:Datsets} reports the dataset statistics, including the explicit OOT check-in ratio $|\vec c_o|/(|\vec c_h|+|\vec c_o|)$.

\paragraph{Data usage protocol.} To make the supervision boundary explicit, Table~\ref{Table:data_protocol} summarizes which part of each user's data is used as \emph{input} versus as a \emph{supervision target} across the three stages. Critically, at test time the destination trajectory is \emph{never} used as input; only the hometown trajectory, the start/end POIs, and the trajectory length (following the standard SPOT-Trip setting) are given. The destination trajectory is used solely for (i) constructing the SFT pseudo-label $\mathbf{y}^{(o)}$ in Stage 1 and (ii) computing training rewards and evaluation metrics.

\begin{table}[!h]
    \centering
    \scriptsize
    \begin{tabular}{p{0.18\columnwidth}|p{0.33\columnwidth}p{0.29\columnwidth}}
    \toprule
    Stage & Input & Target \\
    \midrule
    Stage 1 & H (prompt) & $\mathbf{y}^{(o)}$ from D \\
    Stage 2 & H + Q + masked D & D (masked positions) \\
    Stage 3 & H (prompt) & D (trajectory-level) \\
    Test & H + Q & D (evaluation only) \\
    \bottomrule
    \end{tabular}
    \caption{Data usage protocol. ``H'' means hometown trajectory, ``D'' means destination trajectory, and ``Q'' means destination start/end POIs plus length.}
    \label{Table:data_protocol}
\end{table}

\begin{table*}[!ht]
    \centering
    \scalebox{0.9}{
    \begin{tabular}{l|ccccc}
    \toprule
    Dataset & Users & Regions & POIs & \makecell{Hometown\\check-ins} & \makecell{OOT check-ins\\(ratio)}  \\ \midrule
    Foursquare & 3,007 & 21 & 23,884 & 109,225 & 16,994 (13.46\%)  \\ 
    Yelp & 4,417 & 214 & 29,930 & 58,403 & 20,479 (25.96\%)  \\ 
    \bottomrule
    \end{tabular}
    }
    \caption{Statistics of the two datasets.}
    \label{Table:Datsets}
\end{table*}

\section{Metrics}
\label{Appendix:Metrics}
Let $\hat{V} = [\hat{v}_1, \hat{v}_2, \cdots, \hat{v}_N]$ denote the predicted POI sequence and $V = [v_1, v_2, \cdots, v_N]$ denote the ground-truth POI sequence. Here, $r_{v_i}$ refers to the region of POI $v_i$, and $g_{v_i}$ refers to its category. Both of these sequences exclude the first and last POIs we input into the model. To evaluate the performance of POI sequence prediction, we adopt the following metrics:

1. Hit Rate (HR).
Measures position-wise accuracy by checking whether each predicted POI exactly matches the ground-truth POI at the same index.

\begin{equation}
HR = \frac{1}{N} \sum_{i=1}^{N} \mathbb{I} \big( \hat{v}_i = v_i \big).
\end{equation}

2. Recall Rate (RR).
Evaluates the fraction of unique ground-truth POIs covered by the predictions, regardless of their positions.

\begin{equation}
RR = \frac{ \left| \hat{V} \cap  V\right| }{N}.
\end{equation}

3. Edit Distance (ED).
Computes the normalized Levenshtein edit distance between predicted and ground-truth POI sequences.

4. Dynamic Time Warping Distance (DTW).
Measures sequence alignment cost based on $0/1$ matching, where equality costs $0$ and inequality costs $1$.

\begin{equation}
DTW = \min_{ \pi } \sum_{(i,j) \in \pi} \mathbb{I} \big( \hat{v}_i \neq v_j \big).
\end{equation}

where $\pi$ denotes a warping path aligning $\hat{V}$ and $V$.

5. Category Match (CM).
Measures the consistency in POI categories in the position.

\begin{equation}
CM = \frac{1}{N} \sum_{i=1}^{N} \mathbb{I} \big( g_{\hat{v}_i} = g_{v_i} \big).
\end{equation}

6. Region Match (RM).
Measures the consistency in POI regions in the position.

\begin{equation}
RM = \frac{1}{N} \sum_{i=1}^{N} \mathbb{I} \big( r_{\hat{v}_i} = r_{v_i} \big).
\end{equation}

\section{Implementation Details}
\label{Appendix:Settings}

We set the MAE mask proportion $\rho$ to 0.75, the number of prototypical style embeddings to $M=8$, and $\lambda_D$ to 0.1. The reward weights are $\lambda_{HR}=2$, $\lambda_{RR}=0.5$, $\lambda_{CM}=0.5$, and $\lambda_{DIV}=1$. We train the trajectory model with Adam at a learning rate of $1\times10^{-3}$. Its hidden dimension is 256 and batch size is 4; training stops after 2 epochs without validation improvement. For SemPOI-RL and all LLM-based baselines, we use Qwen3-8B\cite{Qwen3} on four NVIDIA A100 GPUs. We fine-tune the LLM with LoRA rank $r=16$, scaling factor $\alpha=32$, batch size 1 per GPU, 8 gradient-accumulation steps, and a maximum sequence length of 2,048 tokens. Each input contains the user's 50 most recent check-ins. The learning rate is $2\times10^{-5}$ for SFT and $5\times10^{-5}$ for RL; style decoding is capped at 256 tokens.

\begin{figure}
 \centering
  \includegraphics[width=\columnwidth]{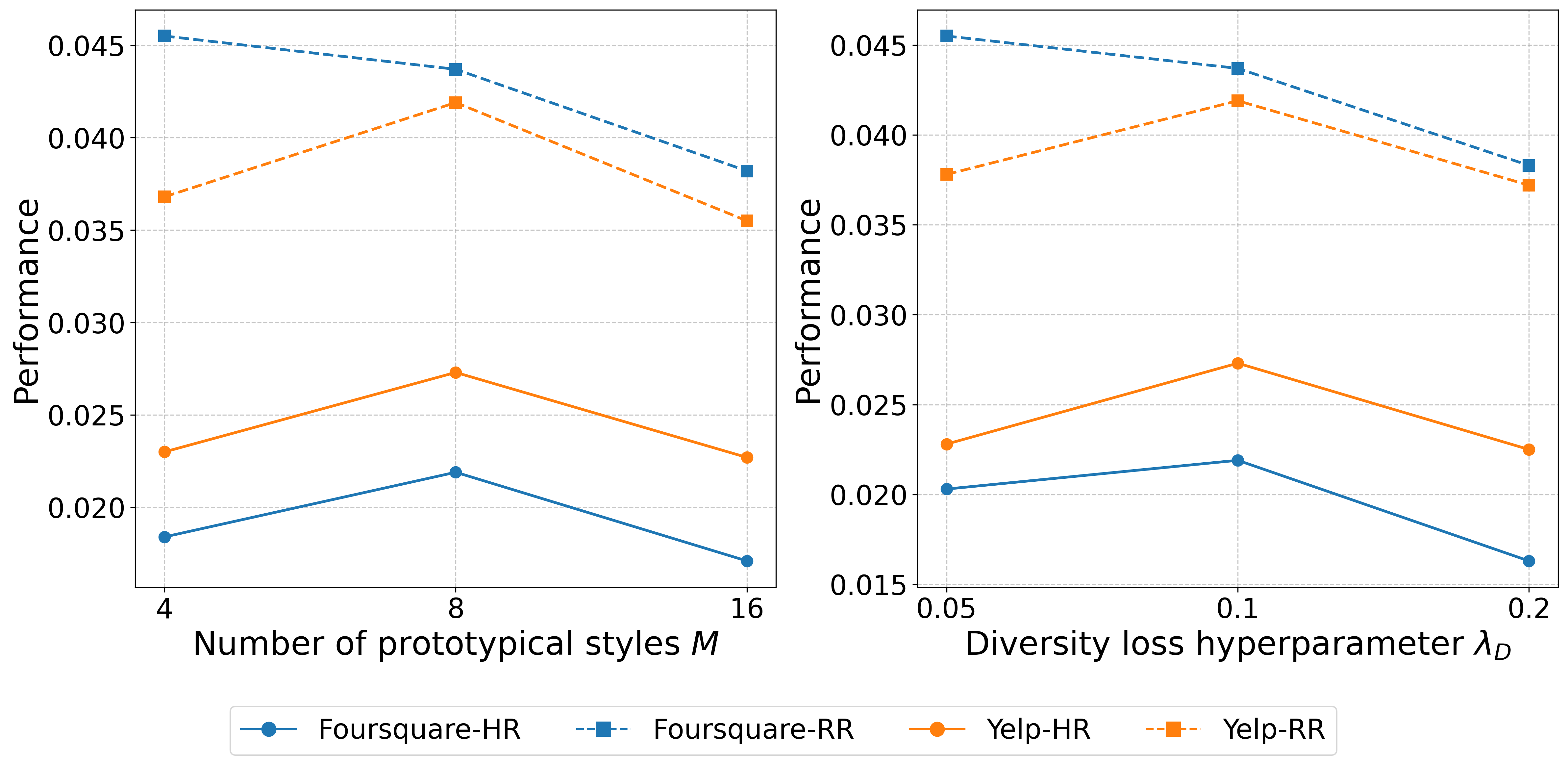}
  \caption{Effects of the number of prototype style embeddings $M$ and diversity regularization coefficient $\lambda_D$ on two datasets.}
  \label{Figure:Hyperparameter}
\end{figure}

\begin{table*}[!h]
\centering
\resizebox{\textwidth}{!}{
\begin{tabular}{l|cccccc|cccccc}
\toprule
& \multicolumn{6}{c|}{Foursquare} & \multicolumn{6}{c}{Yelp} \\
Config & HR & RR & ED & DTW & CM & RM & HR & RR & ED & DTW & CM & RM \\
\midrule
$M=16,\lambda_D=0.1$ & 0.0171 & 0.0382 & 0.9839 & 0.4920 & 0.0322 & \textbf{1.0000} & 0.0227 & 0.0355 & 0.9791 & 0.4896 & 0.0554 & 0.9385 \\
$M=16,\lambda_D=0.2$ & 0.0125 & 0.0370 & 0.9874 & 0.4937 & 0.0356 & \textbf{1.0000} & 0.0226 & 0.0325 & 0.9784 & 0.4892 & 0.0629 & \textbf{0.9578} \\
$M=4,\lambda_D=0.05$ & 0.0187 & 0.0367 & 0.9808 & 0.4904 & 0.0314 & \textbf{1.0000} & 0.0225 & 0.0357 & 0.9780 & 0.4890 & \textbf{0.0706} & 0.9433 \\
$M=4,\lambda_D=0.1$  & 0.0184 & \textbf{0.0455} & 0.9846 & 0.4923 & 0.0364 & \textbf{1.0000} & 0.0230 & 0.0368 & 0.9789 & 0.4893 & 0.0625 & 0.9385 \\
$M=8,\lambda_D=0.05$ & 0.0203 & \textbf{0.0455} & 0.9798 & 0.4889 & 0.0377 & \textbf{1.0000} & 0.0228 & 0.0378 & 0.9792 & 0.4896 & 0.0632 & 0.9522 \\
$M=8,\lambda_D=0.1$  & \textbf{0.0219} & 0.0437 & \textbf{0.9778} & \textbf{0.4876} & \textbf{0.0572} & \textbf{1.0000} & \textbf{0.0273} & \textbf{0.0419} & \textbf{0.9701} & \textbf{0.4843} & 0.0703 & 0.9449 \\
$M=8,\lambda_D=0.2$  & 0.0163 & 0.0383 & 0.9870 & 0.4935 & 0.0349 & \textbf{1.0000} & 0.0225 & 0.0372 & 0.9795 & 0.4898 & 0.0617 & 0.9344 \\
\bottomrule
\end{tabular}}
\caption{Hyper-parameter sensitivity analysis on $M$ and $\lambda_D$. The default setting is $M=8$ and $\lambda_D=0.1$.}
\label{Table:hyper_full}
\end{table*}

\begin{table*}[!h]
\centering
\resizebox{\textwidth}{!}{
\begin{tabular}{l|cccccc|cccccc}
\toprule
& \multicolumn{6}{c|}{Foursquare} & \multicolumn{6}{c}{Yelp} \\
Model & HR & RR & ED & DTW & CM & RM & HR & RR & ED & DTW & CM & RM \\
\midrule
w/o SPAM & 0.0159 & 0.0410 & 0.9838 & 0.4919 & 0.0275 & \textbf{1.0000} & 0.0251 & 0.0387 & 0.9726 & 0.4862 & 0.0580 & 0.9346 \\
w/o RL   & 0.0192 & 0.0422 & 0.9803 & 0.4901 & 0.0552 & \textbf{1.0000} & 0.0227 & 0.0389 & 0.9773 & 0.4887 & 0.0593 & 0.9379 \\
w/o SFT  & 0.0070 & 0.0306 & 0.9930 & 0.4965 & 0.0282 & \textbf{1.0000} & 0.0224 & 0.0366 & 0.9776 & 0.4888 & 0.0569 & 0.9373 \\
SemPOI-RL & \textbf{0.0219} & \textbf{0.0437} & \textbf{0.9778} & \textbf{0.4876} & \textbf{0.0572} & \textbf{1.0000} & \textbf{0.0273} & \textbf{0.0419} & \textbf{0.9701} & \textbf{0.4843} & \textbf{0.0703} & \textbf{0.9449} \\
\bottomrule
\end{tabular}}
\caption{Ablation results on all evaluation metrics.}
\label{Table:ablation_full}
\end{table*}

\begin{table*}[!h]
\centering
\resizebox{\textwidth}{!}{
\begin{tabular}{l|l|cccccc}
\toprule
Setting & Dataset & HR & RR & ED & DTW & CM & RM \\
\midrule
Default $(2:0.5:0.5:1)$ & Foursquare & 0.0219 & 0.0437 & \textbf{0.9778} & \textbf{0.4876} & 0.0572 & \textbf{1.0000} \\
Default $(2:0.5:0.5:1)$ & Yelp      & 0.0273 & 0.0419 & \textbf{0.9701} & \textbf{0.4843} & 0.0703 & \textbf{0.9449} \\
w/o DIV $(2:0.5:0.5:0)$ & Foursquare & \textbf{0.0225} & 0.0431 & 0.9783 & 0.4880 & \textbf{0.0574} & \textbf{1.0000} \\
w/o DIV $(2:0.5:0.5:0)$ & Yelp      & \textbf{0.0279} & 0.0413 & 0.9706 & 0.4847 & \textbf{0.0705} & 0.9445 \\
HR:RR$=1:1$ $(1:1:0.5:1)$ & Foursquare & 0.0213 & \textbf{0.0443} & 0.9781 & 0.4879 & 0.0569 & \textbf{1.0000} \\
HR:RR$=1:1$ $(1:1:0.5:1)$ & Yelp      & 0.0267 & \textbf{0.0425} & 0.9704 & 0.4846 & 0.0700 & 0.9442 \\
\bottomrule
\end{tabular}}
\caption{Effect of RL reward weights. Tuples follow the order (HR, RR, CM, DIV).}
\label{Table:rl_reward}
\end{table*}

\section{Baselines}
\label{Appendix:Baselines}
We compare our proposed framework against several representative baselines:
\noindent
\begin{itemize}[leftmargin=*]
\item LSTM\cite{LSTM}: LSTMs are capable of capturing both short-term and long-term dependencies in sequential patterns, making them more effective for a range of sequential data tasks.
\item GETNext\cite{GETNext}: GETNext is a transformer-based approach that leverages a global trajectory flow map constructed in a user-agnostic manner to enhance next POI prediction. It also introduces a GCN module to generate effective POI embeddings. However, GETNext assumes that all users appearing in the test set are also present in the training set. In contrast, our task focuses on predicting POI sequences for unseen users in out-of-town scenarios, where the test users do not appear in the training data. To adapt GETNext to our setting, we select the user in the training set with the highest Jaccard similarity in POI sequence as the proxy ID for each test user. Moreover, since GETNext is not a sequential generation model, we construct the final predicted POI sequence by selecting the top-$N$ POIs with the highest confidence scores in order.
\item STHGCN\cite{STHGCN}: STHGCN constructs a hypergraph to capture both inter and intra-user relations, and proposes a hypergraph transformer to address the cold-start problem. As its input--output structure is consistent with GETNext, we apply exactly the same experimental adaptation settings when transferring STHGCN to our proposed task.
\item MatTrip\cite{MatTrip}: This method adopts dual LSTM-based encoders to jointly capture users' category-level preferences and the geographical proximity of POIs. An optimized search strategy is further applied to generate user-preferred trip sequences.
\item AR-Trip\cite{AR-Trip}: Built upon a Transformer encoder-only architecture, AR-Trip incorporates prior positional information as auxiliary input to effectively reduce repetitive POIs and recommend more diverse trajectories.
\item KDDC\cite{KDDC}: This approach leverages a knowledge-graph-based representation to strengthen semantic interactions. By modeling the distributions of conformity and individual interests, it aligns the recommended POIs more closely with target user preferences. For fair comparison, we select the top-probability POIs (excluding start and end points) as the final predictions.
\item SPOT-Trip\cite{SPOTTRIP}: SPOT-Trip introduces an ordinary differential equation (ODE) formulation to model the continuous evolution of latent dynamic user preferences. It further incorporates a temporal point process to characterize the instantaneous probability of preference transitions.
\item LLMMove\cite{LLMMove}: LLMMove employs LLMs to extract user preferences and geographical distances via carefully designed prompts. Since our task focuses on out-of-town prediction, we extend the original design by sampling 100 negative candidates in the prompt, allowing the LLM to select the most plausible destination POIs.
\item LLM4POI\cite{LLM4POI}: This SFT-based LLM recommender reformulates the next POI prediction task as a question-answering problem. It introduces prompt-based trajectory similarity to jointly leverage historical trajectories and cross-user trajectories. For our setting, we provide both the user's hometown trajectory and the start--end POIs in the destination city, and directly predict the full trajectory in the destination.
\item Refine-POI\cite{Refine-POI}: A reinforcement learning-based LLM recommender designed for next POI recommendation. It incorporates some recommendation-driven reward functions to enable the LLM to generate effective top-k recommendation lists.
\end{itemize}

\section{Hyper-parameter Analysis}
\label{app:hyper_parameter}
We further investigate the sensitivity of \textbf{SemPOI-RL} to the number of prototype style embeddings $M$, the diversity regularization coefficient $\lambda_D$ in SPAM, and the weight allocation in the RL reward.
Figure~\ref{Figure:Hyperparameter} and Table~\ref{Table:hyper_full} summarize the results. The default setting $M=8$ and $\lambda_D=0.1$ provides the best overall balance across datasets.

When the number of prototype styles is too small, the model cannot adequately represent different temporal phases within a trip. When $M$ is too large, the style space becomes overly fragmented, which harms generalization and weakens alignment quality. This trend is particularly evident on Foursquare, where excessively fine-grained prototypes reduce both HR and RR. This is also reflected in how the prototypes are actually used: we measure prototype usage on the Foursquare test set when $M=16$ and find that usage is highly concentrated, with the top-8 prototypes accounting for about 90\% of the total attention mass while the remaining prototypes are rarely activated. In other words, even when a larger prototype pool is provided, the model effectively relies on only the prototypes it needs through the position-wise attention $\alpha_i$, which can down-weight unused prototypes. Together with the $M$-sensitivity results above, where $M=8$ already yields the best overall balance and larger $M$ over-fragments the style space, this explains why $M=8$ is the preferred setting: it is large enough to cover the active prototypes yet not so large that the style space becomes redundant and harder to align.

The coefficient $\lambda_D$ controls the diversity among prototype style embeddings. A very small $\lambda_D$ leads to homogeneous prototypes, limiting the model's ability to capture stage-specific travel semantics. In contrast, an overly large $\lambda_D$ pushes prototypes to be too dissimilar, which hurts recommendation accuracy. We note that although the position-wise attention can down-weight prototypes that are not needed at a given position, the diversity loss $\mathcal{L}_D$ still pushes all $M$ prototypes apart so that they remain distinct; this is what allows the attention to distribute over genuinely different prototypes rather than homogeneous redundancies. The best trade-off is achieved at $\lambda_D=0.1$.

For the RL reward, our default weight setting is $(2, 0.5, 0.5, 1)$ for \emph{Hit}, \emph{Recall}, \emph{Category}, and \emph{Diversity}, respectively. This design prioritizes correctly positioned POI hits; recall receives a smaller weight, category matching rewards trajectory-level semantic consistency, and diversity prevents repetitive trajectories. The sensitivity results in Table~\ref{Table:rl_reward} show that the model is reasonably robust to moderate changes in reward weights, although different trade-offs may slightly favor HR or RR.

Beyond these hyper-parameters, the Stage-1 prompts enumerate common POI categories (e.g., leisure, nightlife, food exploration, and cultural interest). These examples are retained because they give the LLM a concrete semantic vocabulary for judging which POI categories a user may visit, while the prompt explicitly permits mixed or unlisted styles. We also clarify that the Stage~1 $\rightarrow$ Stage~2 arrow in Figure~\ref{figure:Model} denotes the style embedding $e_o=\mathrm{text\_embedding}(s_o)$ generated by the fine-tuned LLM from the hometown prompt (Eq.~4), rather than the SFT target $\mathbf{y}^{(o)}$.

\section{Statistical Reliability, Efficiency, and Quantitative Interpretability}
\label{app:quant_interp}

This appendix complements the experiments in the main text with (i) a multi-seed statistical-reliability analysis, (ii) an efficiency analysis, and (iii) a set of quantitative interpretability analyses.

\subsection{Statistical Reliability}
\label{app:stat_reliability}
To verify that the improvements are not due to random variation, we run \textbf{SemPOI-RL} and the strongest baseline (\textbf{SPOT-Trip}) over 3 independent random seeds and report the mean$\pm$std of the main metrics in Table~\ref{Table:multiseed}. We also conduct a paired $t$-test between the two methods. The results show that SemPOI-RL is consistently and significantly better than SPOT-Trip on Hit Rate, Recall Rate, Edit Distance, and DTW ($p<0.05$) on both datasets, while the variance across seeds remains small, confirming the stability of the proposed framework.

\begin{table*}[!h]
    \centering
    \footnotesize
    \setlength{\tabcolsep}{3pt}
    \begin{tabular}{l|l|cccc}
    \toprule
    Dataset & Model & HR$\uparrow$ & RR$\uparrow$ & ED$\downarrow$ & DTW$\downarrow$ \\
    \midrule
    \multirow{2}{*}{Foursquare} & SPOT-Trip & 0.0190$\pm$.0010 & 0.0410$\pm$.0014 & 0.9806$\pm$.0011 & 0.4903$\pm$.0009 \\
     & SemPOI-RL & \textbf{0.0219$\pm$.0011}* & \textbf{0.0437$\pm$.0012}* & \textbf{0.9778$\pm$.0010}* & \textbf{0.4876$\pm$.0008}* \\
    \midrule
    \multirow{2}{*}{Yelp} & SPOT-Trip & 0.0248$\pm$.0012 & 0.0389$\pm$.0016 & 0.9726$\pm$.0012 & 0.4863$\pm$.0010 \\
     & SemPOI-RL & \textbf{0.0273$\pm$.0013}* & \textbf{0.0419$\pm$.0015}* & \textbf{0.9701$\pm$.0011}* & \textbf{0.4843$\pm$.0009}* \\
    \bottomrule
    \end{tabular}
    \caption{Multi-seed ($n{=}3$) mean$\pm$std of SemPOI-RL and the strongest baseline. Bold indicates the better mean; * marks a statistically significant difference (paired $t$-test, $p<0.05$).}
    \label{Table:multiseed}
\end{table*}

\subsection{Efficiency Analysis}
\label{app:efficiency}
Table~\ref{Table:efficiency} reports the computational cost on four NVIDIA A100 GPUs. At inference time, the per-user cost is dominated by at most 256-token style generation plus one MAE forward pass; the full test set finishes within 1 hour. Directly placing tens of thousands of POIs in an LLM prompt is infeasible, so the LLM baselines use a simulated pre-ranked set of 100 destination candidates. This protocol still assumes a strong coarse-ranking stage, whereas SemPOI-RL's MAE scores the destination catalog directly. Our pipeline is slower than simple recommenders, but its measured cost is moderate relative to the accuracy and interpretability gains.

\begin{table}[!h]
    \centering
    \footnotesize
    \begin{tabular}{l|c}
    \toprule
    Component & Wall-clock time \\
    \midrule
    LLM supervised fine-tuning & 1 hour \\
    SPAM + MAE training & 5 hours \\
    GRPO RL & 5 hours \\
    Full-test-set inference & $<1$ hour \\
    \bottomrule
    \end{tabular}
    \caption{Wall-clock cost on four NVIDIA A100 GPUs.}
    \label{Table:efficiency}
\end{table}

\subsection{Quantitative Interpretability Analysis}
\label{app:quant_interp_analysis}
Beyond the qualitative case study in the main text, we provide two quantitative analyses to assess interpretability.

\paragraph{(1) Style fidelity via text similarity.}
We embed the inferred destination style $s_o$ and the ground-truth destination style summary $\mathbf{y}^{(o)}$ (the supervision target used in Stage-1 SFT) with \textbf{Qwen3-Embedding-4B}\cite{Qwen3Embedding} and report their average cosine similarity. As shown in Table~\ref{Table:style_sim}, similarity increases monotonically from the pre-SFT model to the SFT model and finally to the RL-tuned model, on both datasets. This confirms that SFT teaches the LLM to produce destination-oriented styles, and RL further sharpens them toward the ground-truth travel style.

\begin{table}[!h]
    \centering
    \footnotesize
    \begin{tabular}{l|cc}
    \toprule
    Stage & Foursquare & Yelp \\
    \midrule
    Before SFT & 0.6326$\pm$0.0718 & 0.6200$\pm$0.1166 \\
    After SFT  & 0.6623$\pm$0.0895 & 0.6813$\pm$0.0971 \\
    After RL   & \textbf{0.6751$\pm$0.0693} & \textbf{0.6940$\pm$0.0985} \\
    \bottomrule
\end{tabular}
    \caption{Average cosine similarity (mean$\pm$std) between the inferred destination style $s_o$ and the ground-truth style summary $\mathbf{y}^{(o)}$, measured by Qwen3-Embedding-4B.}
    \label{Table:style_sim}
\end{table}

\paragraph{(2) Prototype-to-category statistics over the full test set.}
Since prototypes are latent vectors and cannot be directly converted to readable text, we characterize each prototype indirectly by the distribution of the 12 coarse-grained POI categories it activates, computed separately for the \emph{ground-truth} and the \emph{predicted} trajectories. The categories are: Food \& Dining, Travel \& Transport, Other, Entertainment \& Nightlife, Shopping, Outdoor \& Nature, Culture \& History, Sports \& Fitness, Health \& Wellness, Work \& Business, Home \& Living, and Education. Table~\ref{Table:proto_cat} reports the top-3 categories for the 4 main prototypes under SFT+RL on the Foursquare test set. The predicted prototype-category distributions are broadly consistent with the ground-truth distributions, supporting our (scaled-down) interpretability claim that the prototypes provide \emph{approximate, category-mediated, trip-phase-specific} attribution.

\begin{table*}[!h]
    \centering
    \footnotesize
    \begin{tabular}{c|c|c}
    \toprule
    Prototype & Ground-truth Top-3 & Predicted Top-3 \\
    \midrule
    P1 & Food \& Dining, Other, Travel \& Transport & Other, Travel \& Transport, Food \& Dining \\
    P2 & Travel \& Transport, Other, Food \& Dining & Other, Travel \& Transport, Entertainment \& Nightlife \\
    P3 & Food \& Dining, Other, Entertainment \& Nightlife & Food \& Dining, Other, Entertainment \& Nightlife \\
    P4 & Food \& Dining, Other, Travel \& Transport & Other, Travel \& Transport, Shopping \\
    \bottomrule
    \end{tabular}
    \caption{Top-3 activated POI categories per prototype (SFT+RL, Foursquare test set), for the ground-truth vs. the predicted trajectory.}
    \label{Table:proto_cat}
\end{table*}

\section{Final Output Examples}
\label{app:final_examples}

\paragraph{Multi-sample comparison before and after RL.}
We further examine individual users whose style fidelity and prediction accuracy both improve after RL, and report the full generated style text for two representative Foursquare cases in Tables~\ref{Table:case_781} and~\ref{Table:case_27}. The ground-truth style is produced by an initial LLM from the observed destination trajectory; the three generated variants are obtained before SFT (Base), after SFT, and after RL, respectively.

For uid=781, the Base style is qualitatively close but generic. SFT raises similarity from 0.613 to 0.654, yet its wording still drifts toward hometown routines by mentioning gyms and offices, and its trajectory Hit drops to 0.143. RL refocuses the style on destination nightlife and dining, raises similarity to 0.727, and restores Hit to 0.286 (Recall 0.400). For uid=27, the Base style is biased toward the hometown trajectory; SFT aligns it toward a cultural and social characterization; and RL further sharpens the university, pub-based nightlife, and boutique/bagel dining cues, increasing similarity from 0.613 to 0.637 to 0.651 and Recall from 0.500 to 0.667.

\begin{table*}[!h]
\centering
\footnotesize
\renewcommand{\arraystretch}{1.15}
\begin{tabular}{p{0.12\textwidth}p{0.82\textwidth}}
\toprule
Stage (Sim) & Travel Style \\
\midrule
Ground-truth & The user's travel style in Los Angeles reflects a blend of leisure, nightlife, and cultural exploration, with a clear interest in social and recreational experiences. The frequent visits to gay bars suggest a preference for vibrant nightlife and social interaction, while the stay at hotels on consecutive nights indicates a possible business or extended leisure trip. The inclusion of a Mexican restaurant highlights an interest in food exploration and local cuisine. The visit to a monument and scenic lookout points to a cultural and outdoor-oriented mindset, showing an appreciation for both historical landmarks and natural beauty. The trip appears to be a mix of relaxation and discovery, with a focus on enjoying the city's diverse offerings, from nightlife to scenic spots. The user seems to prioritize experiences that combine social engagement, cultural enrichment, and leisurely exploration, indicating a dynamic and adventurous travel approach. \\
Base (0.613) & The user exhibits a travel style characterized by a blend of social engagement, cultural exploration, and urban discovery. They frequently visit bars and nightlife spots, suggesting an interest in socializing and experiencing vibrant evening scenes. Their visits to a variety of restaurants, including Mexican, American, and burger joints, indicate a preference for diverse culinary experiences. The inclusion of movie theaters, malls, and clothing stores points to an appreciation for entertainment, shopping, and retail environments. Additionally, their visits to gyms and offices suggest a balance between personal health and professional commitments. The user also shows an interest in niche venues like gay bars and thrift stores, indicating openness to unique and eclectic experiences. Overall, they likely enjoy a mix of leisure, social interaction, and exploration in a new city, favoring urban areas with a dynamic nightlife, varied dining options, and cultural attractions. \\
After SFT (0.654) & The user's travel style in the hometown trajectory suggests a pattern of routine and leisurely exploration, with a strong emphasis on social and recreational activities. The frequent visits to bars and restaurants, especially late at night, indicate a preference for nightlife and social interaction. The presence of gyms and offices points to a balance between personal health and professional commitments. The user also shows an interest in dining experiences, visiting a variety of restaurant categories, including Mexican, American, and burger joints, suggesting a taste for diverse culinary options. The occasional visits to cultural and entertainment venues like movie theaters and malls imply an appreciation for leisure and shopping. Overall, the travel behavior reflects a lifestyle that values social engagement, relaxation, and a mix of work and play, with a clear interest in exploring different aspects of urban life. \\
After RL (0.727) & The user exhibits a pattern of travel that suggests a blend of leisure and social exploration, with a strong interest in nightlife and dining experiences. The frequent visits to bars, especially in the evenings and early mornings, indicate a preference for socializing and nightlife activities. The user also shows an appreciation for diverse food experiences, visiting a variety of restaurant types such as Mexican, American, and burger joints, which suggests a curiosity for different cuisines. Additionally, the presence of visits to cultural and entertainment venues like movie theaters and a tea room hints at an interest in leisure and cultural activities. The user's travel style appears to be relaxed and exploratory, with a focus on personal enjoyment and social interaction rather than business or productivity-oriented activities. The occasional visits to shopping areas and clothing stores suggest a mild interest in retail and urban exploration. Overall, the travel behavior reflects a lifestyle centered around social engagement, food, and entertainment. \\
\bottomrule
\end{tabular}
\caption{Full travel-style text before and after RL for uid=781 (Los Angeles; Hit 0.286, Recall 0.400). ``Sim'' is the cosine similarity between the generated style and the ground-truth style (Qwen3-Embedding-4B).}
\label{Table:case_781}
\end{table*}

\begin{table*}[!h]
\centering
\footnotesize
\renewcommand{\arraystretch}{1.15}
\begin{tabular}{p{0.12\textwidth}p{0.82\textwidth}}
\toprule
Stage (Sim) & Travel Style \\
\midrule
Ground-truth & The user's travel style in Austin reflects a strong interest in cultural and entertainment experiences, with a focus on music and nightlife. The sequence of visits suggests a preference for social and leisure activities, particularly in venues that offer live music, such as rock clubs and music venues. The user also shows an appreciation for local food, as evidenced by the visit to a BBQ joint. The timing of these visits, especially in the evenings and on weekends, indicates a tendency to engage in nightlife and entertainment. Additionally, the visit to a convention center and a hotel hints at possible business or event-related travel, though the overall pattern leans more toward leisure and cultural exploration. The user appears to be someone who enjoys immersive experiences in Austin's vibrant music scene and is likely drawn to the city's dynamic social and artistic environment. This travel behavior suggests a preference for active, engaging, and culturally rich experiences rather than passive or routine activities. \\
Base (0.613) & The user's hometown trajectory suggests a preference for a mix of cultural, social, and experiential activities, with a focus on urban exploration and engagement with diverse environments. The inclusion of a university, pub, boutique, bagel shop, performing arts venue, and tech startup indicates an interest in both intellectual and social settings, as well as unique and locally curated experiences. This pattern implies a traveler who enjoys discovering new places, engaging with local culture, and seeking out both casual and curated experiences. In a new city, the user is likely to prioritize areas with a vibrant arts scene, diverse dining options, and opportunities for social interaction, while also showing an openness to exploring innovative or niche businesses. The user may appreciate a balance between leisure and discovery, favoring destinations that offer both cultural depth and a lively atmosphere. \\
After SFT (0.637) & The user's travel style in New York suggests a blend of cultural engagement and leisurely exploration, with a clear interest in both social and artistic experiences. The visit to a university indicates a possible academic or intellectual curiosity, while the subsequent stop at a pub reflects a preference for social interaction and nightlife. The presence of multiple visits to a bagel shop on the same day highlights a fondness for local cuisine and possibly a relaxed, exploratory approach to dining. The visit to a performing arts venue further reinforces an appreciation for cultural and artistic activities, suggesting a traveler who values enriching experiences and is open to discovering new aspects of a city. Overall, the trajectory reflects a balanced mix of social, cultural, and culinary interests, indicating a traveler who enjoys both exploration and meaningful engagement with the local environment. \\
After RL (0.651) & The user's travel style in New York suggests a preference for a mix of cultural and social experiences, with a clear interest in both leisure and exploration. The visit to a university indicates a potential interest in knowledge or academic environments, while the subsequent stop at a pub points to a social and possibly nightlife-oriented behavior. The boutique and bagel shop visits suggest an appreciation for local culture, unique shopping, and casual dining. The performing arts venue visit highlights a strong interest in cultural and artistic activities, and the tech startup visit implies a curiosity about innovation and business environments. Overall, the user appears to enjoy a balanced mix of cultural engagement, social interaction, and exploration of urban spaces, reflecting a dynamic and curiosity-driven travel style. \\
\bottomrule
\end{tabular}
\caption{Full travel-style text before and after RL for uid=27 (Austin; Hit 0.333, Recall 0.667). ``Sim'' is the cosine similarity between the generated style and the ground-truth style (Qwen3-Embedding-4B).}
\label{Table:case_27}
\end{table*}

\clearpage
\raggedbottom

\section{Prompt Templates for Travel Style Generation}
\label{app:prompts}
In this appendix, we provide the two prompt templates used in Stage 1. The first prompt asks the LLM to summarize a user's destination-oriented travel style from the observed destination trajectory. The second prompt asks the LLM to infer the user's likely destination travel style from the hometown trajectory. In both prompts, the region name is represented by \texttt{\{region\_name\}} and the trajectory text is represented by \texttt{\{trajectory\_text\}}.

\subsection{Prompt for Summarizing Destination Travel Style}
\label{app:prompt_destination}

\begin{lstlisting}[
language={},
basicstyle=\ttfamily\small,
breaklines=true,
breakindent=0pt,
breakautoindent=false,
columns=fullflexible,
keepspaces=true,
xleftmargin=0pt,
framexleftmargin=0pt
]
You are an expert in user travel behavior analysis.

A user has visited the following points of interest in {region_name}.

Destination trajectory:
{trajectory_text}

Please summarize the user's travel style in this region based on the trajectory within 200 words.
Your summary should describe the user's overall preferences, such as activity patterns, lifestyle tendencies, and possible interests reflected by the visited places.

Requirements:
1. Write one coherent paragraph in natural language.
2. Focus on high-level travel style rather than listing POIs one by one.
3. Capture semantic preferences such as leisure, nightlife, food exploration, cultural interest, business activity, outdoor preference, or shopping tendency.
4. Do not mention that you are an AI model.
5. Do not output bullet points or JSON.
6. Keep the summary concise but informative.

Travel style summary:
\end{lstlisting}

\subsection{Prompt for Inferring Destination Travel Style from Hometown Trajectory}
\label{app:prompt_hometown}

\begin{lstlisting}[
language={},
basicstyle=\ttfamily\small,
breaklines=true,
breakindent=0pt,
breakautoindent=false,
columns=fullflexible,
keepspaces=true,
xleftmargin=0pt,
framexleftmargin=0pt
]
You are an expert in user travel behavior analysis.

A user has the following historical trajectory in their hometown region {region_name}.

Hometown trajectory:
{trajectory_text}

Based on this hometown trajectory, infer what travel style the user is likely to exhibit when visiting another city within 200 words. Your goal is not to describe the hometown itself, but to infer the user's destination-oriented travel preferences in an out-of-town scenario.

Requirements:
1. Write one coherent paragraph in natural language.
2. Infer high-level travel style and likely preferences in a new city.
3. Focus on transferable interests, such as whether the user may prefer restaurants, nightlife, cultural attractions, shopping areas, parks, business-related places, or mixed urban exploration.
4. Do not simply restate the visited POIs; provide an abstract style-level inference.
5. Do not mention that you are an AI model.
6. Do not output bullet points or JSON.
7. Keep the summary concise but informative.

Inferred destination travel style:
\end{lstlisting}

\end{document}